\documentclass{bmvc2k}
\usepackage{amssymb}
\usepackage{tabularx, booktabs}
\usepackage{floatrow}
\usepackage{caption}
\usepackage{multirow}
\usepackage{bm}

\newcolumntype{Y}{>{\centering\arraybackslash}X}

\title{Learning Fine-Grained Visual Understanding for Video Question Answering via \\ Decoupling Spatial-Temporal Modeling}

\addauthor{Hsin-Ying Lee}{shinlee@cmlab.csie.ntu.edu.tw}{1}
\addauthor{Hung-Ting Su}{}{1}
\addauthor{Bing-Chen Tsai}{}{1}
\addauthor{Tsung-Han Wu}{}{1}
\addauthor{Jia-Fong Yeh}{}{1}
\addauthor{Winston H. Hsu}{}{1,2}

\addinstitution{
 National Taiwan University
}
\addinstitution{Mobile Drive Technology}

\runninghead{Lee et al.}{Fine-grained Understanding for Video Question Answering}

\def\eg{\emph{e.g}\bmvaOneDot}

\begin{document}

\maketitle

\begin{abstract}
While recent large-scale video-language pre-training made great progress in video question answering, the design of spatial modeling of video-language models is less fine-grained than that of image-language models; existing practices of temporal modeling also suffer from weak and noisy alignment between modalities. To learn fine-grained visual understanding, we decouple spatial-temporal modeling and propose a hybrid pipeline, Decoupled Spatial-Temporal Encoders, integrating an image- and a video-language encoder. The former encodes spatial semantics from larger but sparsely sampled frames independently of time, while the latter models temporal dynamics at lower spatial but higher temporal resolution. To help the video-language model learn temporal relations for video QA, we propose a novel pre-training objective, Temporal Referring Modeling, which requires the model to identify temporal positions of events in video sequences. Extensive experiments demonstrate that our model outperforms previous work pre-trained on orders of magnitude larger datasets. Our code is available at \url{https://github.com/shinying/dest}.
\end{abstract}


\section{Introduction}

Videos are the complex composition of human actions, objects, scenes, and their interactions over time. To examine the capability of machines for video understanding, video question answering (video QA), a task of answering questions about videos, is proposed and requires machines to associate questions in natural languages with visual contents, including scenes \cite{xu2017video, Yu_2018_ECCV}, dialogues \cite{lei2018tvqa, Choi_On_Heo_Seo_Jang_Lee_Zhang_2021}, temporal relationships \cite{Jang_2017_CVPR, Yu_Xu_Yu_Yu_Zhao_Zhuang_Tao_2019, GrundeMcLaughlin2021AGQA, xiao2021next}, and higher-order cognition \cite{lei-etal-2020-likely,Yi2020CLEVRER, xiao2021next}. Recent breakthroughs were achieved by pre-training a deep multi-modality encoder, mostly Transformer \cite{Vaswani2017}, with large-scale video-language datasets \cite{Miech_2019_ICCV, yang2021just, Bain21}. Models first learned semantic connections between visual and linguistic contents and then were fine-tuned on downstream video-language tasks \cite{li-etal-2020-hero, Zhu_2020_CVPR, yang2021just, zellersluhessel2021merlot, Seo_2022_CVPR}.

Despite the advance of this framework in video QA, the spatial semantics encoding of video-language (VL) models is not as fine-grained as the sophisticated design for image-language (IL) models \cite{Anderson_2018_CVPR,Zhang_2021_CVPR,seo2021look}.
A preliminary analysis shows that on video QA benchmarks entailing spatial and temporal knowledge, simply averaging frame-by-frame predictions of an IL model can sometimes outperform state-of-the-art VL models.
Though the VL models exhibit a slight advantage in questions involving temporal information, the IL model greatly excels in capturing spatial clues (improvement by 7\% accuracy; see the full results in Section \ref{sec:exp:pre:spa}). 
The positive performance of IL models could also be attributed to the nature of video QA: the answers to the questions pertaining to only spatial semantics, without specifying time, are usually consistent across all related frames.
This property suggests the potential of encoding fine-grained spatial semantics with only IL models.

\begin{figure}[t]
\centering
\includegraphics[width=\linewidth]{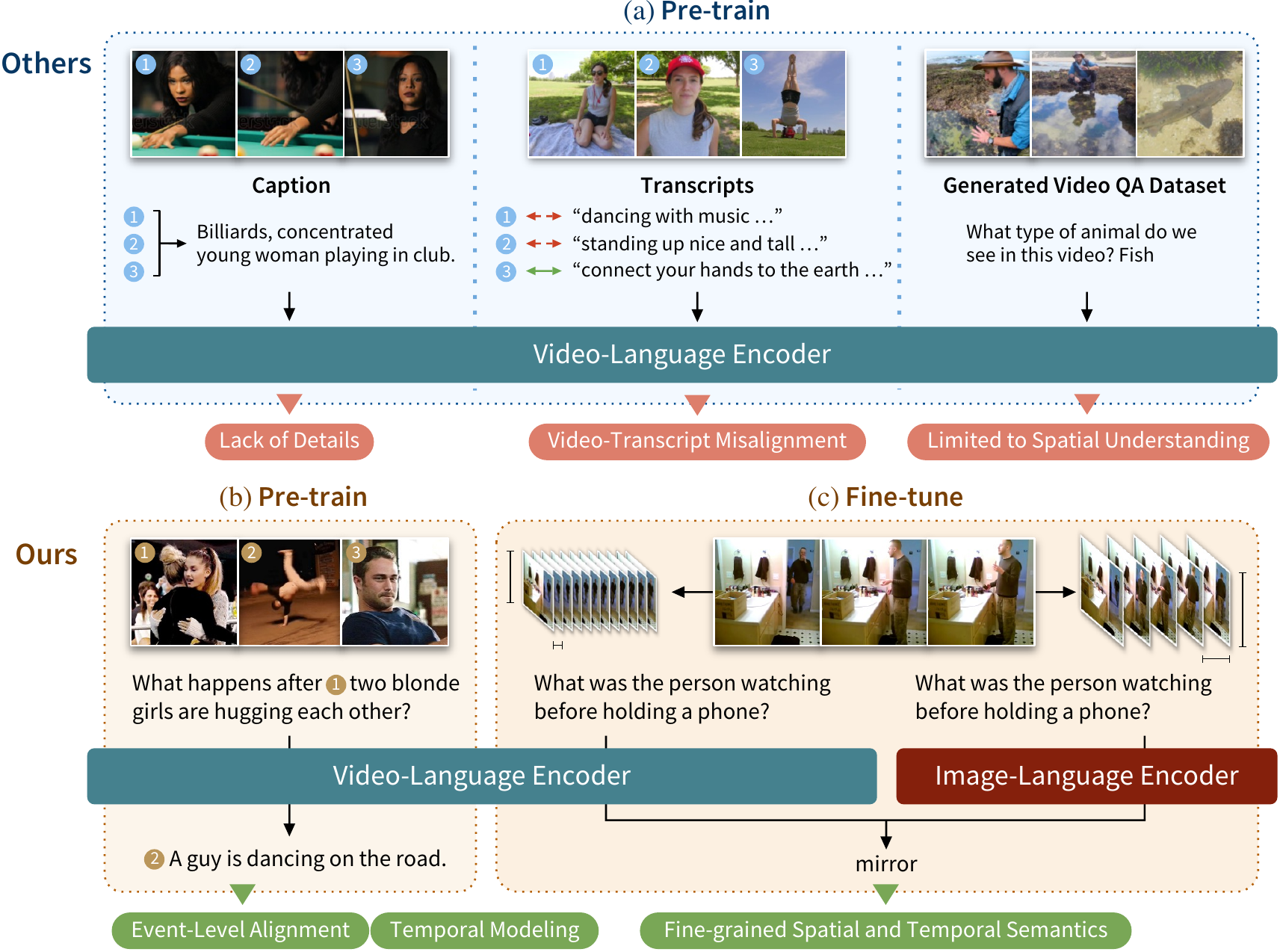}
\caption{\textcolor{bmv@captioncolor}{Comparison between (a) previous and (b)(c) our approaches for video QA.} (a) Prior work solved video QA by video-language pre-training but might suffer from lack of event details, video-transcript misalignment or limited diversity of pre-training questions. (b) We pre-train a video-language encoder to learn event representations and temporal relations between them by asking the model to identify specific events in synthesized video sequences. (c) We integrate the video-language model with a pre-trained image-language model to encode fine-grained spatial and temporal semantics at different spatial-temporal resolutions.}

\label{fig:teaser}
\end{figure}

In addition to spatial modeling, prior work modeled only coarse-grained temporal relations. A question involving temporal relations in video QA often refers to specific events happening in periods of time and inquires about the order of events \cite{Jang_2017_CVPR, Yu_Xu_Yu_Yu_Zhao_Zhuang_Tao_2019, GrundeMcLaughlin2021AGQA, xiao2021next}. It is thus essential to model events in videos and associate the sequence with time conjunctions in questions, such as \textit{before} and \textit{after}. However, as the examples in Figure \ref{fig:teaser} (a), prior approaches \cite{Zhu_2020_CVPR, seo2021look, violet_2022, Seo_2022_CVPR, all_in_one_2022} aligning a video with a sentence might lose details of sequential events (what happens after the woman hit the ball), while matching short clips with transcripts \cite{li-etal-2020-hero, zellersluhessel2021merlot} may suffer from noise as spoken words often contain something not related to scenes \cite{miech20endtoend}. Others \cite{yang2021just, yang2022learningta} pre-training on generated video QA datasets were mostly limited to spatial understanding. In fact, another examination reveals that the performance with shuffled frame inputs of some of these approaches is similar to that with normal inputs on video QA benchmarks requiring temporal modeling (see more details in Section \ref{sec:exp:pre:tem}). The result suggests developing a more effective strategy for modeling temporal relations.

To obtain fine-grained encoding of spatial and temporal semantics for video QA, we propose a novel pipeline, Decoupled Spatial-Temporal Encoders (DeST), decoupling spatial-temporal modeling into IL and VL encoders, illustrated in Figure \ref{fig:teaser} (c). With IL models well-versed in fine-grained spatial modeling, we incorporate a pre-trained IL model to encode static spatial information independent of time from sparsely sampled frames at high spatial resolution. For questions requiring temporal relations, we train a VL encoder to model temporal dynamics, operating at high temporal but low spatial resolution. These two streams complement each other by paying attention to disparate aspects of videos.

To effectively model temporal relations for video QA, the VL encoder has to recognize events in videos, build their temporal relations, and associate such relations with languages containing temporal information. Thus, we introduce a novel pre-training objective, Temporal Referring Modeling (TRM). Depicted in Figure \ref{fig:teaser} (b), TRM queries absolute and relative positions of events in videos synthesized by concatenating clips sampled from video captioning datasets \cite{tgif-cvpr2016, Wang_2019_ICCV}. The concatenation simulates transitions of scenes and events in videos. Answering such queries requires a model to aggregate contiguous frames into events and distinguish adjacent events from distant ones. These operations help a model learn both short- and long-term temporal dynamics.

We validate our model on two video QA benchmarks, ActivityNet-QA \cite{Yu_Xu_Yu_Yu_Zhao_Zhuang_Tao_2019} and AGQA 2.0  \cite{grunde2022agqa}. The former contains diverse question types requiring spatial or temporal semantics, and the latter weaves spatial and temporal information together in each question to evaluate compositional reasoning. DeST outperforms the previous state-of-the-art. The ablation studies also demonstrate the efficacy of the proposed pipeline and pre-training objective.

In summary, we make the following key contributions. (\emph{i}) With IL and VL models demonstrating complementary advantages, we decouple spatial and temporal modeling into a hybrid pipeline composed of both models to encode fine-grained visual semantics. (\emph{ii}) We present a novel pre-training objective, Temporal Referring Modeling, to learn temporal relations between events by requiring models to identify specific events in video sequences. (\emph{iii}) We outperform previous VL state-of-the-art methods on two benchmarks with orders of magnitudes less data for pre-training.
\vspace{-4pt}
\section {Related Work}

\vspace{-6pt}
\subsection{Video Question Answering}

To encode, accumulate and build relationships between visual contents and between modalities for video QA, conventional approaches adopted Recurrent Neural Networks with attention \cite{Zeng_Chen_Chuang_Liao_Niebles_Sun_2017, xu2017video, zhao2017video, zhao2017video-1, Jang_2017_CVPR}, Memory Networks \cite{tapaswi2016movieqa, na2017read, Gao_2018_CVPR, Kim_2019_CVPR, Fan_2019_CVPR}, Graph Neural Networks \cite{huang2020location, Jiang_Han_2020, Park_2021_CVPR, Liu_2021_ICCV, peng2021progressive, xiao2022video}, Modular Networks \cite{LeMinh2021}, and self-attention \cite{Li_Song_Gao_Liu_Huang_He_Gan_2019, Jiang_Chen_Lin_Zhao_Gao_2020, urooj-etal-2020-mmft}. By pre-training large-scale VL datasets, Transformers \cite{Vaswani2017} have further improved the interaction between modalities and made great progress in video QA \cite{Zhu_2020_CVPR, violet_2022, seo2021look, all_in_one_2022, li-etal-2020-hero, zellersluhessel2021merlot, yang2021just, yang2022learningta}. Our approach is built on the benefit of modeling relationships with pre-trained Transformers. In contrast to prior work, we carefully examine and take the individual advantage of IL and VL pre-training to encode spatial and temporal semantics. 

\vspace{-8pt}
\subsection{Pre-training for Temporal Relation Modeling}\label{sec:rel:pretrain}

VL pre-training learns to model temporal relationships via different approaches.

\noindent
\textbf{Learning from Global Alignment.} 
\cite{Sun_2019_ICCV, Zhu_2020_CVPR, seo2021look, violet_2022, all_in_one_2022, luo2020univl} pre-trained models on datasets where a sentence delineates a single event of the entire corresponding video.
With features of two modalities being aligned globally, events happening sequentially in a video are compressed, and details of events not mentioned in descriptions are likely lost. Such representations are not fine-grained enough for questions referring to specific moments.

\noindent
\textbf{Learning from Local Alignment and Frame Ordering.}
\cite{li-etal-2020-hero, zellersluhessel2021merlot} pre-trained models over datasets with dense annotations such as video transcripts \cite{Miech_2019_ICCV}. They matched segmented visual features with utterances and required models to order shuffled or any two frames. With this approach, models learn event-level but weak alignment between videos and languages as spoken words do not always correspond to visual contents \cite{miech20endtoend}.  
Besides, ordering frames without grounding in languages makes models learn, instead of temporal relations, rational predictions of what is likely to happen before and after an event, which is more related to visual common sense \cite{agrawal-etal-2016-sort, park2020visualcomet, Hwang2021COMETATOMIC2O}.

\noindent
\textbf{Learning from Large-Scale Video Question Answering Datasets.}
\cite{yang2021just, yang2022learningta} pre-trained VL models over large-scale video QA datasets. The diversity of pre-training questions thus determines the effectiveness and capacity of transferred knowledge, but generated questions in \cite{yang2021just} and \cite{yang2022learningta} mainly pertain to scene and dialogue understanding, leaving temporal relationship modeling unsolved.

\vspace{-8pt}
\subsection{Encoding Motion and Appearance}

Prior arts have explored two-stream networks to encode motion and appearance for action recognition \cite{NIPS2014_00ec53c4, NIPS2016_3e7e0224, Feichtenhofer_2016_CVPR, wang_2016_temporal, Feichtenhofer_2017_CVPR, Diba_2017_CVPR}. \cite{Feichtenhofer_2019_ICCV,diba_2020_large} combined different spatial and temporal resolution to separately encode slow- and fast-changing scenes, and
\cite{Ryoo2020AssembleNet, ryoo_2020_assemblenet++} searched for multi-stream connectivity. Analogously, our two streams complement each other by focusing on disparate aspects of videos, but while their two streams both encode short-term actions, our IL stream aggregates scene information independent of time,
and the VL stream encodes entire videos and constructs the temporal relationships between all actions and events.

Some recent work revealed that understanding temporality is not always necessary to solve VL tasks. \cite{Lei_2021_CVPR,lei_2022_revealing} taking sparsely sampled frames outperformed previous methods. \cite{Buch_2022_CVPR} provided stronger baselines with single frame inputs. However, with new tasks requiring temporal modeling proposed, such conclusions are likely to be circumscribed. We thus take a further step by proposing an effective strategy to encode fine-grained temporal semantics.
\vspace{-16pt}
\section {Method}

We introduce our video QA pipeline, Decoupled Spatial-Temporal Encoders (Section \ref{sec:method:model}), and the pre-training objective, Temporal Referring Modeling (Section \ref{sec:method:trm}). Implementation details are described in the supplement (Section \ref{sec:impl}).

\subsection{Decoupled Spatial-Temporal Encoders}\label{sec:method:model}

The coarse-grained spatial modeling of prior approaches motivates us to develop more effective architectures, and IL models have shown great potential. While most VL models take scene or multi-frame features pre-extracted by image or action recognition models \cite{luo2020univl,li-etal-2020-hero,zellersluhessel2021merlot,yang2021just}, region features \cite{lu2019vilbert, tan-bansal-2019-lxmert, Su2020VL-BERT,Zhang_2021_CVPR} and features processed by attention \cite{xu2015show, Anderson_2018_CVPR} have been proved powerful for IL models. These features provide detailed information about visual elements along with their spatial relations. Since static scene information, if asked by questions without specifying time, are usually consistent across related frames, IL models should also be competent to encode fine-grained spatial relations for video QA. 

Hence, we propose Decoupled Spatial-Temporal Encoders (DeST), a video QA pipeline decoupling spatial and temporal modeling into an IL and a VL encoder. The IL encoder takes unordered and sparsely sampled frames at high spatial resolution as input. Fine-grained spatial information of static scenes is obtained by building a consensus among these frames. The VL encoder with input action features at high temporal resolution recognizes and models the transitions of actions and events. These two streams of information are fused at the final stage to jointly form the prediction. We leave other ways of fusion for future exploration.

\begin{figure}
\RawFloats
\begin{minipage}{.56\textwidth} \centering
\includegraphics[height=5.7cm]{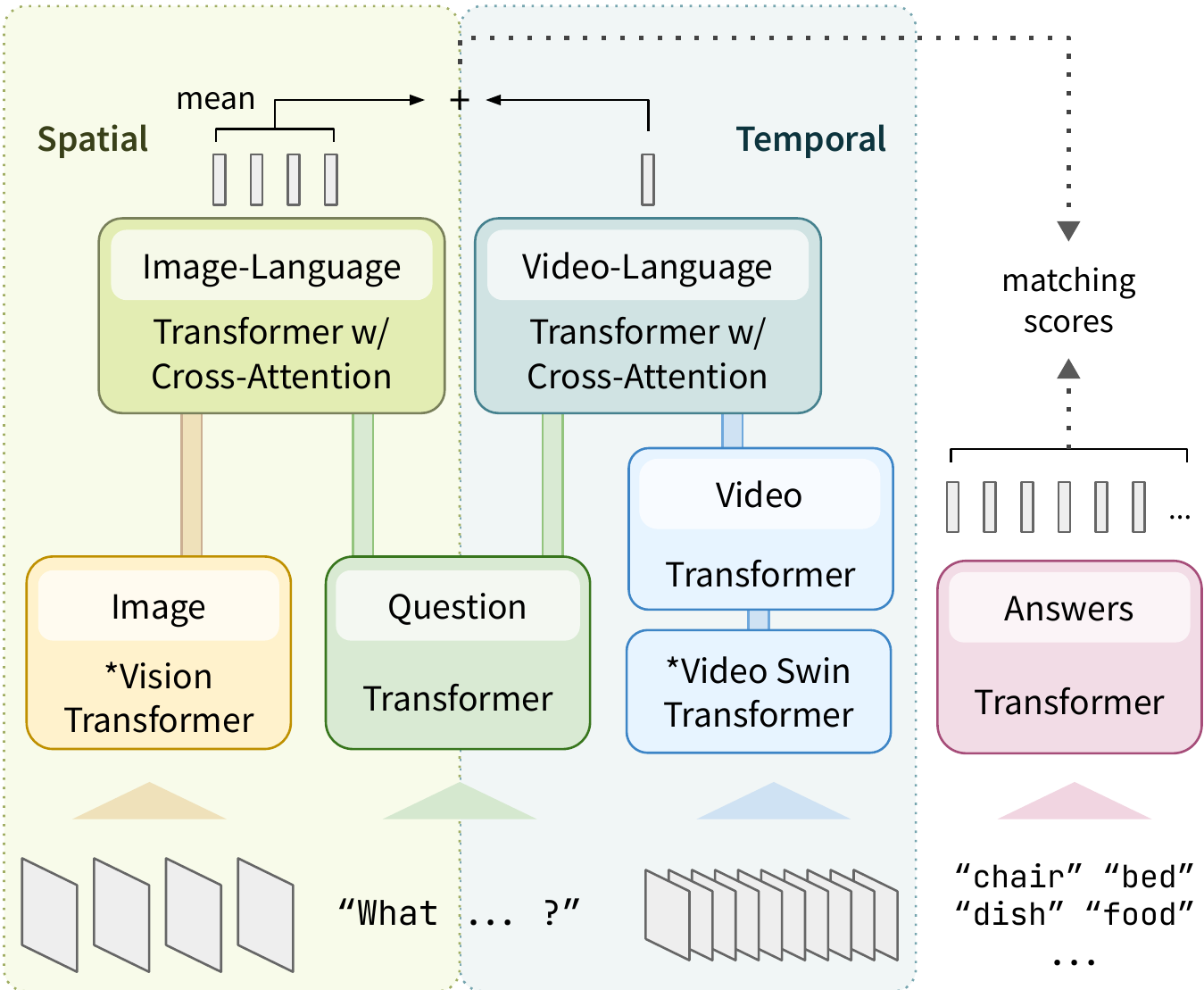}
\vspace{12pt}
\captionof{figure}{\textcolor{bmv@captioncolor}{Decoupled Spatial-Temporal Encoders.} Encoded questions are fused with frames and videos to gather spatial and temporal information. Their representations are then compared with answer candidates to obtain the final predictions. (* marks the frozen modules.)}
\label{fig:pipeline}
\end{minipage} \hspace{0.03\textwidth}
\begin{minipage}{.40\textwidth} \centering
\includegraphics[height=5.6cm]{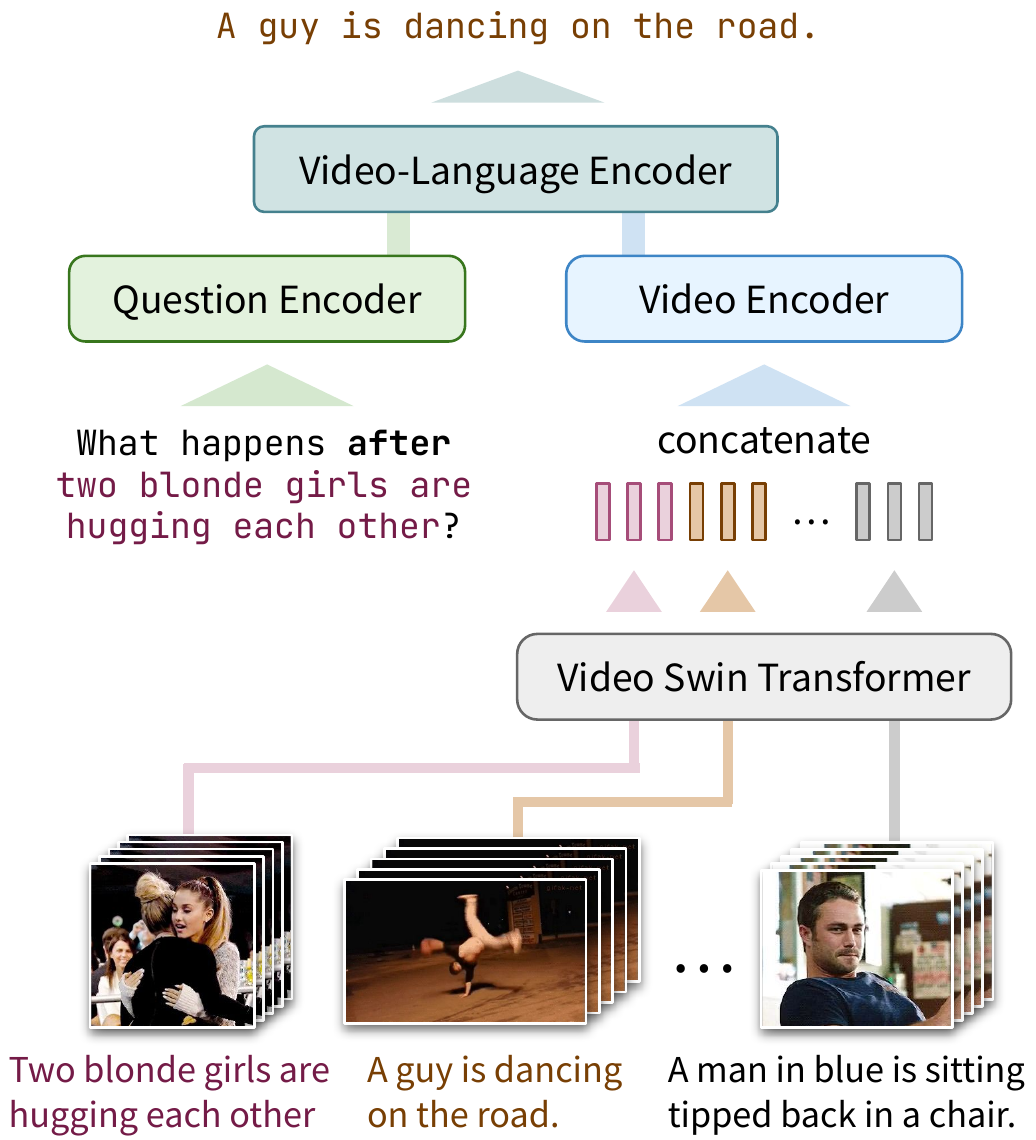}
\vspace{0pt}
\captionof{figure}{\textcolor{bmv@captioncolor}{Temporal Referring Modeling,} which associates visual events and their temporal relationships with languages by asking absolute and relative positions of events in concatenated video features sampled from video captioning data.}
\label{fig:trm}
\end{minipage}
\end{figure}

As illustrated in Figure \ref{fig:pipeline}, DeST consists of an image encoder, a video encoder, and a question encoder to process inputs, as well as an IL encoder and a VL encoder, both with cross-attention \cite{li2021align, jaegle2021perceiver, li2022blip, lei_2022_revealing}, to perform multi-modality interaction. Another answer encoder encodes answer candidates, similar to \cite{yang2021just}. To answer a question about a video, the question, video, and frames that are sparsely sampled from the video are encoded by their respective encoders. The question features then perform cross-attention to both frame and video features. The sum of two multi-modality representations is finally compared with encoded answer candidates to obtain the prediction. Formally, $\mathcal{Q}$ denotes the input question. $\{\mathcal{I}^1, ..., \mathcal{I}^T\}$ are $T$ frames sampled from the input video $\mathcal{V}$, where $T \ll$ the length of $\mathcal{V}$. The question $\mathcal{Q}$ is first encoded into a sequence of embeddings $\mathbf{w} = \{w_\mathtt{cls}, w_1, ..., w_L\},\ w \in \mathbb{R}^D$, where $w_\mathtt{cls}$ is the embedding of the \texttt{[CLS]} token, and $L$ is the number of word tokens. Then $\mathbf{w}$ is fused with the frames and video as described below.

\noindent
\textbf{Image-Language Encoding.} For each $t$ from 1 to $T$, the image encoder transforms frame $\mathcal{I}^t$ into a sequence of patch embeddings $\mathbf{u} = \{u^t_\mathtt{cls}, u^t_1, ..., u^t_N\},\ u \in \mathbb{R}^D$, where $N$ is the number of patches. Then the question feature $\mathbf{w}$ and frame feature $\mathbf{u}$ are fused by the IL encoder with cross-attention and transform into $\{x^t_\mathtt{cls}, x^t_1, ..., x^t_L\},\ x \in \mathbb{R}^D$. The multi-modality representation of the IL stream $r$ is the average of \texttt{[CLS]} token embeddings $x^t_\mathtt{cls}$ of all frames encoded by a final multi-layer perceptron (MLP):
\begin{equation}
    r = \frac{1}{T}\sum_{t=1}^{T} \text{MLP} (x^t_\mathtt{cls}), \ r \in \mathbb{R}^D.
\end{equation}

\noindent
\textbf{Video-Language Encoding.}
The video feature extractor first encodes the input video $\mathcal{V}$ into a sequence of features $\mathbf{e} = \{e_1, ..., e_M\},\ e \in \mathbb{R}^H$, where $M$ is the length of the feature sequence. To indicate the beginning and the end of the video, we add two learnable tokens before and after the feature sequence. Temporal position encoding is also added to each feature to indicate the temporal order. Next, the feature sequence $\mathbf{e}$ are contextualized and transformed into $\mathbf{v} = \{v_\mathtt{bos}, v_1, ..., v_M, v_\mathtt{eos}\},\ v \in \mathbb{R}^D$, where $v_\mathtt{bos}$ and $v_\mathtt{eos}$ are the beginning and the end token after contextualization. The question feature $\mathbf{w}$ then performs cross attention to the video feature $\mathbf{v}$ through the VL encoder and transforms into $\{y_\mathtt{cls}, y_1, ..., y_L\},\ y \in \mathbb{R}^D$. The multi-modality representation of the VL stream $s \in \mathbb{R}^D$ is the output of the first token $y_{\mathtt{cls}}$ transformed by a final MLP.

\noindent
\textbf{Answer Selection.} Following \cite{yang2021just}, another text encoder encodes the answer candidates (collected from all answers in training data with frequency $> 1$ for open-ended QA). The prediction of each candidate is the dot product between each encoded candidate and the sum of two multi-modality representations. Formally, $\mathcal{A}$ denotes the answer set. For all $a \in \mathcal{A}$, we take the \texttt{[CLS]} token $z^a_\mathtt{cls} \in \mathbb{R}^D$ of $a$'s feature. Then the logit of $a$ is obtained via:
\begin{equation}
    p_a = (r + s)^\mathsf{T}z^a_\mathtt{cls},\ p \in \mathbb{R}.
\end{equation}

\subsection{Temporal Referring Modeling}\label{sec:method:trm}

To pre-train the multi-modality encoders with affordable computation resources, we adopt an IL encoder pre-trained with image question answering (image QA), specifically VQA \cite{balanced_vqa_v2}, and train the VL encoder for fine-grained temporal modeling with a novel objective.

Modeling fine-grained temporal relations for video QA requires the encoder to understand videos as event sequences and to associate the temporal relations of events with descriptions containing time conjunctions. Therefore, we develop Temporal Referring Modeling (TRM), which, in the form of video QA, inquires about absolute and relative temporal positions of events in videos. As depicted in Figure \ref{fig:trm}, given a video composed of multiple events, TRM asks the model four questions: what happens at the beginning, at the end, before an event, or after an event? The model then selects an event description as the answer. To accomplish this task requires the model to identify events and manage the order.

TRM needs VL data that offers (1) event-level annotations that delineate scenes and events for segments of videos and (2) descriptions that explain the temporal dynamics of these segments. Dense video captioning \cite{krishna2017dense} should be ideally suited for our needs, but unfortunately, many of its time segments overlap, making the temporal relations ambiguous, and labeling cost also hinders scalability. To satisfy the two conditions, we thus develop a simple yet effective way to generate data. As the example in Figure \ref{fig:trm}, we concatenate videos sampled from video captioning datasets to create videos with scene and event transitions. Then we generate questions by completing templates with captions of these videos. Incorrect answers are the other captions in the same video sequences, making the task more difficult.

Take, as an example, generating a video and a question that asks which event happens after an event. We first sample $K$ pairs from a video captioning dataset, with each pair $k$ composed of a video $\mathcal{V}_k$ and a caption $\mathcal{C}_k$. The videos are encoded by the feature extractor into feature sequences $\{e_1^k, ..., e_{M_k}^k\}$ for all $k$ from 1 to $K$, where $M_k$ is the length of features of $\mathcal{V}_k$. These sequences are then concatenated and form $\mathbf{e} = \{e_1^1, ..., e_{M_1}^1, e_1^2, ..., e_{M_K}^K\}$. To generate the question, we first sample a captions $\mathcal{C}_i$ where $1 \leq i < K,\ i \in \mathbb{N}$. Then the question $\mathcal{Q}$ is ``What happens after $\mathcal{C}_i$?'' with the choices $\mathcal{A} = \{\mathcal{C}_k\ |\ 1 \leq k \leq K,\ k \neq i,\ k \in \mathbb{N}\}$ and the correct answer $\mathcal{C}_{i+1}$. Other questions are constructed similarly, where the answers to the questions about the beginning and the end are $\mathcal{C}_1$ and $\mathcal{C}_K$ respectively. With all input the same as general video QA, the encoded feature $\mathbf{w}$ of question $\mathcal{Q}$ and the video feature $\mathbf{e}$ are input to the VL encoder, going through the encoding and contextualizing process described in Section \ref{sec:method:model}. The final objective is to minimize a standard cross-entropy loss.
\section {Experiments}

We elaborate on the preliminary analysis of spatial and temporal reasoning capability of prior work (Section \ref{sec:exp:pre}). Then we demonstrate the improvement in two video QA benchmarks with DeST and TRM (Section \ref{sec:exp:trm}). The ablation studies are lastly presented evaluating the efficacy of each component. (Section \ref{sec:exp:abl}).

\begin{table}[t]
\RawFloats
\centering \footnotesize \setlength{\tabcolsep}{3pt}
\begin{minipage}{.49\textwidth} \centering
    \begin{tabular}{lccccc}
        \toprule
        Type            & Just-Ask & VIOLET & ALBEF & UB \\
        \midrule
        Motion          & 28.00 & 18.25 & 32.50 & ~~70.63   \\
        Spatial Rel.    & 17.50 & 15.00 & 24.38 & ~~75.63   \\
        Temporal Rel.   & ~~4.88 & ~~2.12 & ~~3.75 & ~~32.88 \\
        Yes / No        & 66.28 & 71.87 & 79.75 & 100.00    \\
        Color           & 34.29 & 31.28 & 57.39 & ~~98.99   \\
        Object          & 26.73 & 22.33 & 31.45 & ~~70.13   \\
        Location        & 35.75 & 30.57 & 36.01 & ~~86.79   \\
        Number          & 50.17 & 50.33 & 55.61 & ~~99.83   \\
        Other           & 36.82 & 33.02 & 40.16 & ~~71.98   \\
        \midrule
        Overall         & 38.86 & 37.44 & 46.66 & ~~80.74   \\
        \bottomrule
    \end{tabular}
    \vspace{12pt}
    \captionof{table}{\textcolor{bmv@captioncolor}{Comparison between prior methods and our upper bound of ActivityNet-QA by question type.} ALBEF exhibits advantages on the questions involving spatial reasoning. (Rel. is short for Relationships, and UB is the abbreviation for upper bound.)}
    \label{tab:pre-spa}
\end{minipage} \hspace{.02\textwidth}
\begin{minipage}{.45\textwidth} \centering
    \begin{tabular}{ccl}
        \toprule
        Method & Benchmark & Accuracy  \\
        \midrule
        \multirow{2}{*}{VIOLET} & AGQA~~ & 49.15 \\
        & AGQA* & 49.22{\scriptsize$\pm$.02}  \\
        \midrule
        \multirow{2}{*}{Just-Ask} & AGQA~~ & 51.27 \\ 
        & AGQA* & 47.73{\scriptsize$\pm$.06} \\
        \midrule
        \multirow{2}{*}{HERO} & VIOLIN~~ & 69.01 \\ 
        & VIOLIN* & 68.71{\scriptsize$\pm$.08} \\
        \bottomrule
    \end{tabular}
    \vspace{12pt}
    \captionof{table}{\textcolor{bmv@captioncolor}{Results of prior work taking shuffled frames as input}. The little performance drop indicates that some methods are not sensitive to the order of frames. (* signifies that input frames are shuffled. We report the average of three results for the shuffle experiment.)}
    \label{tab:pre-tem}
\end{minipage}

\end{table}
\subsection{Preliminary Analysis}\label{sec:exp:pre}

\noindent
\textbf{Baselines.}
We take ALBEF \cite{li2021align} as an example of IL models. For VL models, we study VIOLET \cite{violet_2022}, HERO \cite{li-etal-2020-hero}, and Just-Ask \cite{yang2021just}, which respectively instantiate three approaches discussed in Section \ref{sec:rel:pretrain}. These are state-of-the-art of each approach with public code bases.

\subsubsection{Encoding Spatial Semantics}\label{sec:exp:pre:spa}

We first assess the ability of encoding spatial semantics of IL models and VL models\footnote{Just-Ask and VIOLET as HERO does not support open-ended QA}. ALBEF is run as image QA by sampling frames from a video and averaging frame predictions.

\noindent
\textbf{Benchmark.} 
We conduct the analysis on ActivietNet-QA \cite{Yu_Xu_Yu_Yu_Zhao_Zhuang_Tao_2019}, which contains 5.8K videos of human activities in daily life and 58K question-answer pairs spanning diverse categories across spatial and temporal semantics offering comprehensive evaluations.

\noindent
\textbf{Results.}
Table \ref{tab:pre-spa} contrasts the accuracy (acc) by question type of the IL model with other VL models. ALBEF, though without temporal modeling, is adept at spatial reasoning, such as \emph{Spatial Relationships} and \emph{Color}, while Just-Ask demonstrates a slight advantage in \emph{Temporal Relationships}. Due to the removal of rare answers following \cite{yang2021just}, we report our performance upper bound of each type, which is the proportion of questions in the test set whose answers appeared in the training set. The tiny number of \emph{Temporal Relationships} reveals the long-tailed distribution of its answers, which partially explains the poor performance.

\begin{table}[t]
\RawFloats
\centering \footnotesize \setlength{\tabcolsep}{3pt}
\begin{minipage}{.53\textwidth} \centering
    \begin{tabular}{llc}
        \toprule
        Method & Pre-training Data & Acc \\
        \midrule
        CoMVT \cite{seo2021look} & 100M vid & 38.8 \\
        Just-Ask \cite{yang2021just} & 69M vid & 38.9   \\
        MV-GPT \cite{Seo_2022_CVPR} & 100M vid & 39.1 \\
        SiaSamRea \cite{NEURIPS2021_dea18482}  & 5.6M img  & 39.8 \\
        MERLOT \cite{zellersluhessel2021merlot} & 180M vid & 41.4   \\
        VIOLET \cite{violet_2022} & 180M vid + 2.5M vid + 3M img & 37.5 \\
        FrozenBiLM \cite{yang2022zero} & 10M vid & 43.2 \\
        Singularity \cite{lei_2022_revealing} & 14M img + 2.5M vid & 44.1 \\
        \midrule
        DeST (ours)        & 14M img + 120K VQA + 14K vid & \textbf{46.8} \\
        \bottomrule
    \end{tabular}
    \vspace{12pt}
    \captionof{table}{\textcolor{bmv@captioncolor}{Comparison with previous methods on ActivityNet-QA.} We outperform all methods with significantly less pre-training data. The dataset names are provided in the supplement Section \ref{sec:dataset}. (img: images. vid: videos.)}
    \label{tab:anet}
\end{minipage} \hspace{.03\textwidth}
\begin{minipage}{.35\textwidth} \centering
    \begin{tabular}{lccc}
        \toprule
        Type      & Best  & DeST  & Diff (\%)  \\
        \midrule
        Motion          & 32.50 & 35.75 & \textbf{10.00} \\
        Spatial Rel.    & 24.38 & 23.88 & ~-2.05 \\
        Temporal Rel.   & ~~4.88 & ~~5.25 & ~~\textbf{7.58} \\
        Yes / No        & 79.75 & 78.61 & ~-1.43 \\
        Color           & 57.39 & 59.11 & ~~3.00 \\
        Object          & 31.45 & 30.50 & ~-3.02 \\
        Location        & 36.01 & 36.27 & ~~0.72 \\
        Number          & 55.61 & 55.28 & ~-0.59 \\
        Other           & 40.16 & 39.63 & ~-1.32 \\
        \midrule
        Overall         & 46.66 & \textbf{46.79} & ~~0.28 \\
        \bottomrule
    \end{tabular}
    \vspace{12pt}
    \captionof{table}{\textcolor{bmv@captioncolor}{Comparison with prior methods on AcivityNet-QA by question type}. We perform comparably in question types of spatial information and improve temporal modeling.}
    \label{tab:anet-type}
\end{minipage}
\end{table}

\subsubsection{Modeling Temporal Relationships}\label{sec:exp:pre:tem}

We evaluate the capability of modeling temporal relationships by shuffling input frames and measuring the performance drop. Models are first trained with normal input and tested their performance with shuffled input. Intuitively, taking shuffled frames as input should be detrimental to the performance of the questions requiring temporal modeling, such as those inquiring about the order of actions or events in videos.

\noindent
\textbf{Benchmarks.}
For VIOLET and Just-Ask, we conduct the study on AGQA 2.0 \cite{grunde2022agqa}, a large-scale open-ended video QA benchmark where spatial and temporal information is required in each question for evaluating compositional reasoning. 
It contains 2.27M question-answer pairs and 9.6K videos. For HERO, we consider VIOLIN \cite{Liu_2020_CVPR}, a task of judging hypotheses from visual premises, which has been officially tested in their experiments.

\noindent
\textbf{Result.}
In Table \ref{tab:pre-tem}, Just-Ask demonstrates the slight capability of temporal modeling, while VIOLET and HERO are not sensitive to the order of input frames, and their performances of taking normal and shuffled input frames are similar. The result suggests clear insufficiency for temporal relationship modeling.

\setbox0\hbox{\scriptsize \setlength{\tabcolsep}{0pt} \tabular{@{}l}Reasoning\endtabular}
\setbox1\hbox{\scriptsize \setlength{\tabcolsep}{0pt} \tabular{@{}l}Semantic\endtabular}
\setbox2\hbox{\scriptsize \setlength{\tabcolsep}{0pt} \tabular{@{}l}Structure\endtabular}
\setbox3\hbox{\scriptsize \setlength{\tabcolsep}{0pt} \tabular{@{}l}Overall\endtabular}

\begin{table}[t]
\RawFloats
\centering \footnotesize \setlength{\tabcolsep}{3pt}
\begin{minipage}{.55\textwidth} \centering
    \begin{tabular}{clccc}
        \toprule
        & Type            & Best w/o PT  & Best w/ PT & DeST  \\
        \midrule
        \multirow{8}{*}{\rotatebox{90}{\usebox0}}
        & Object-Relationship   & 40.33 & 48.91 & \textbf{59.66}  \\
        & Relationship-Action   & 49.95 & 66.55 & \textbf{72.98}  \\
        & Object-Action   & 50.00 & 68.78 & \textbf{75.20}  \\
        & Superlative     & 33.55 & 39.83 & \textbf{48.94}  \\
        & Sequencing      & 49.78 & 67.01 & \textbf{73.53}  \\
        & Exists          & 50.01 & 59.35 & \textbf{63.21}  \\
        & Duration Comparison   & 47.03 & 50.49 & \textbf{60.39}  \\
        & Activity Recognition  & ~~5.52  & 21.53 & \textbf{27.78}  \\
        \midrule
        \multirow{3}{*}{\rotatebox{90}{\usebox1}}
        & Object          & 40.40 & 49.31 & \textbf{61.27}  \\
        & Relationship    & 49.99 & 59.60 & \textbf{63.93}  \\
        & Action          & 47.58 & 58.03 & \textbf{65.96}  \\
        \midrule
        \multirow{5}{*}{\rotatebox{90}{\usebox2}}
        & Query           & 36.34 & 47.98 & \textbf{61.22}  \\
        & Compare         & 49.71 & 65.11 & \textbf{72.04}  \\
        & Choose          & 46.56 & 46.90 & \textbf{53.01}  \\
        & Logic           & 50.02 & 56.20 & \textbf{59.18}  \\
        & Verify          & 50.01 & 58.13 & \textbf{63.02}  \\
        \midrule
        \multirow{3}{*}{\rotatebox{90}{\usebox3}}
        & Binary          & 48.91 & 55.35 & \textbf{62.61}  \\
        & Open            & 36.34 & 47.98 & \textbf{61.22}  \\
        & All             & 42.11 & 51.27 & \textbf{61.91}  \\
        \bottomrule
    \end{tabular}
    \vspace{12pt}
    \captionof{table}{\textcolor{bmv@captioncolor}{Comparison with prior work on AGQA 2.0.} We list the best performance among methods without (Best w/o PT) and with pre-training (Best w/ PT) for each question type. DeST exceeds all methods in all question types.}
    \label{tab:agqa}
\end{minipage}\hspace{.04\textwidth}
\begin{minipage}{.36\textwidth} \centering
    \begin{tabular}{cccc}
        \toprule
        Question & Frames & Video & Acc \\
        \midrule
        $\checkmark$ &         &         & 41.32 \\
        $\checkmark$ & $\checkmark$ &         & 50.07 \\
        $\checkmark$ & VQA     &         & 51.00 \\
        $\checkmark$ &         & $\checkmark$ & 51.08 \\
        $\checkmark$ &         & TRM     & 55.62 \\
        $\checkmark$ & VQA     & $\checkmark$ & 56.61 \\
        $\checkmark$ & VQA     & TRM*    & 56.97 \\
        \midrule
        $\checkmark$ & VQA     & TRM     & 61.91 \\
        \bottomrule
    \end{tabular}
    \vspace{12pt}
    \captionof{table}{\textcolor{bmv@captioncolor}{Ablation study of input modalities and pre-training strategies on AGQA 2.0.} The results favor our hybrid pipeline and TRM. ($\checkmark$ means the modality is presented. VQA: pretrained on VQA. TRM: pre-trained with TRM. *: shuffled input.)}
    \label{tab:ablate}
    \vspace{6pt}
    \begin{tabular}{lc}
        \toprule
        Training Stream & Acc \\
        \midrule
        Image-Language  & 49.91 \\
        Video-Language  & 16.56 \\
        \midrule
        Both            & 61.91 \\
        \bottomrule
    \end{tabular}
    \vspace{12pt}
    \captionof{table}{\textcolor{bmv@captioncolor}{Ablation study of two encoding streams on AGQA 2.0.}}
    \label{tab:not-ensem}
\end{minipage}

\end{table}
\subsection{Video Question Answering}\label{sec:exp:trm}

DeST takes frames and videos as input. Frames are extracted at 3 FPS, following \cite{lei2018tvqa}; then we sample $T$ frames randomly during training and uniformly during inference, similar to the strategy for action recognition. Video features are also pre-extracted by the video encoder and excluded from the optimization. More details are left in the supplement (Section \ref{sec:model-arch}).

Table \ref{tab:anet} compares DeST with prior work on ActivityNet-QA. We outperform all previous methods with orders of magnitudes less pre-training data. The performance of each question type is listed in Table \ref{tab:anet-type}, where \emph{Best} shows the highest scores among the three methods in Table \ref{tab:pre-spa}. This rigorous comparison leads to a more comprehensive analysis in terms of both spatial and temporal modeling. \emph{Diff} lists the difference between \emph{Best} and our performance in proportion to \emph{Best}. Our hybrid model performs, as expected, comparably with the IL model in spatial modeling since we are not improving IL processing. On the other hand, the performance of categories such as \emph{Motion} and \emph{Temporal Relationships} are boosted, verifying the efficacy of TRM.

Table \ref{tab:agqa} presents the performance on AGQA 2.0, which offers extensive annotation of multiple abilities necessary to answer each question. We list the highest accuracy among the methods without pre-training reported by \cite{grunde2022agqa} (Best w/o PT) and the higher scores between our implementation of Just-Ask and VIOLET (Best w/ PT). DeST surpasses all prior work in all question types. Besides, while TRM is similar to only the questions of \emph{Sequencing}, which accounts for about 7\% of the dataset, TRM can serve as an abstraction of temporal modeling and generalize to other question types, such as \emph{Relationship-Action} and \emph{Object-Action}, which inquire about the temporal relationship between human actions and their interactions with objects. The full table and detailed analysis are provided in the supplement (Section \ref{sec:agqa-full}).

\subsection{Ablation Studies}\label{sec:exp:abl}

We present the influence of input modalities and pre-training over AGQA 2.0 to study the effect of modeling decisions. As presented in Table \ref{tab:ablate}, question-only input reveals the language bias, which serves as a baseline. The boost in performance with frames and videos suggests successful encoding. Pretraining the IL encoder with VQA and the VL encoder with TRM both enhance the modeling capacity further. The performance drop due to shuffling videos verifies the efficacy of TRM. The full results are included in the supplement (Section \ref{sec:agqa-full}).

In Table \ref{tab:not-ensem}, we ablate the IL or VL stream. A model is trained with both streams and tested on AGQA 2.0 with a single stream. The performance drastically drops in both settings, proving that our hybrid model is not a trivial ensemble. It might also be noted that the overwhelming advantage of the IL stream over its VL counterpart cannot conclude the utility of any stream, for each stream can be trained to perform better than the question-only baseline. We hypothesize that temporal information can be seen as the complex evolution of spatial information, and thus when both streams collaborate in spatial-temporal modeling, the IL stream offers an overall understanding of visual elements and scenes, while the VL stream assists it and models the detailed changes.
\section {Conclusion}

In this work, considering the complementary advantage of image- and video-language models, we decouple spatial-temporal modeling and propose a hybrid pipeline for video QA, where an image-language encoder encodes spatial information and a video-language encoder models temporal dynamics. To capture event-level temporal relations, the video-language encoder is pre-trained with an objective to identify events in videos by their temporal positions. With the collaboration between image- and video-language models as well as fine-grained temporal modeling, we advance the visual understanding for video QA.
\section*{Acknowledgement}

This work was supported in part by the National Science and Technology Council under Grant MOST 110-2634-F-002-051, Mobile Drive Technology Co., Ltd (MobileDrive), and NOVATEK fellowship. We are also grateful to the National Center for High-performance Computing.

\bibliography{ref}

\clearpage

\appendix

\title{Supplementary Material}
\maketitlesup

In this supplement, we provide additional and clarifying details for the main paper. Section \ref{sec:impl} contains implementation details including the model architecture, pre-training objectives, datasets, parameters of optimization, and computational cost of our model. Section \ref{sec:exps} expands the experimental results of Table \ref{tab:pre-tem}, \ref{tab:ablate}, and \ref{tab:not-ensem} in the main paper and offers the analysis of the model behavior in different question types. We also conduct additional experiments testing the modeling decision on ActivityNet-QA, as well as evaluating the influence of temporal resolutions of the image-language model, the number of concatenated videos for Temporal Referring Modeling, and the loss combination strategy.
\section{Implementation Details}\label{sec:impl}

\subsection{Model Architectures}\label{sec:model-arch}

We introduce the details of our Decoupled Spatial-Temporal Encoders (DeST). Following \cite{li2021align} and \cite{violet_2022}, the image encoder is a 12-layer Vision Transformer \cite{dosovitskiy2021an}, and the video encoder contains a Video Swin Transformer \cite{Liu_2022_CVPR} (Swin-B) pre-trained on Kinetics-600 \cite{kinetics} for feature extraction and a 6-layer Transformer for contextualization. The question and answer encoder are both 6-layer Transformers \cite{Vaswani2017} with each layer composed of a self-attention operation and a feed-forward network (FFN). The image- and video-language encoder are two 6-layer Transformers where each layer contains an additional cross-attention operation \cite{li2021align, jaegle2021perceiver, jaegle2022perceiver, li2022blip, lei_2022_revealing}, in which text features serve as queries and perform attention to visual features. The question, image, and image-language encoder are the same as the modules of ALBEF \cite{li2021align} pre-trained on VQA \cite{balanced_vqa_v2}. The video contextualization module and video-language encoder are initialized from the question and image-language encoder respectively. The image and video encoder are fixed during the whole training process. The detailed parameters are listed in Table \ref{tab:param_arch}.
\begin{table}[h]
\centering \small
    \begin{tabular}{lc}
    \toprule
        Hyperparameter & Value   \\
    \midrule
        Embedding Size ($D$)        & 768  \\
        Number of Patches ($N$)     & 576  \\
        Video Feature Size ($H$)    & 1024 \\
        FFN Inner Hidden Size       & 3072 \\
        Number of Attention Heads   & 12   \\
        Attention Dropout           & 0.1  \\
        Dropout                     & 0.1  \\
    \bottomrule
    \end{tabular}
    \caption{Hyperparameters for the architecture.}
    \label{tab:param_arch}
\end{table}

Since the optimization of video encoding is not included in video-language training, we extract and store video features to save memory. We operate the Video Swin Transformer with the same configuration as Swin-B, which samples every two frames and transforms a window of 32 frames into one feature. For long videos, such as ActivityNet \cite{Yu_Xu_Yu_Yu_Zhao_Zhuang_Tao_2019} with an average length of 180 seconds, we shift the window by 32 frames. For others, such as the datasets used in pre-training or AGQA 2.0 \cite{grunde2022agqa}, we shift the window by 16 frames, and thus every window overlaps with half of its previous and next window. Features of extremely long videos are sampled such that all videos are within a limited length.

\subsection{Video-Language Pre-training}

\subsubsection{Details of Question and Video Synthesis for Temporal Referring Modeling}
Temporal Referring Modeling (TRM) generates questions to inquire about absolute and relative temporal positions of specific events in videos. Questions are formed by choosing from five templates and filling in the templates with video descriptions. The choice of templates includes ``What happens?'', ``What happens at the beginning?'', ``What happens at the end?'', ``What happens before \texttt{[event x]}?'', and ``What happens after \texttt{[event x]}?'', where the first question is irrelevant to temporal relations but incorporated to facilitate video-language matching. The other four questions are designed for resemblance to video QA requiring temporal modeling, such as \emph{Temporal Relationships} in ActivityNet-QA \cite{Yu_Xu_Yu_Yu_Zhao_Zhuang_Tao_2019} or \emph{State Transition} in TGIF-QA \cite{Jang_2017_CVPR}.

Except for the first question paired with a single video, the corresponding videos of other questions are synthesized by concatenating videos sampled from video captioning datasets. This operation simulates a sequence of events that happen one after another and provides us with the exact position of each event. 

One may be concerned that the transitions of events in real videos are rather smooth and ambiguous, instead of clear differences between videos in a random concatenated video sequence, where people, objects, and almost the entire scenes drastically change. For example, in a video where people clean up the table after finishing dinner in the dining room, most of the visual elements, such as the people and furniture, remain the same, but we humans can easily recognize these two events by comparing the actions and interactions between the people in the video. While TRM cannot generate such videos, our model has learned a similar capability with TRM to compare human actions and interactions between moments. During fine-tuning, it can focus on adapting to smooth transitions and thus learn faster than models with neither the capability of temporal reasoning nor event recognition.

\subsubsection{Auxiliary Objective with Contrastive Learning}\label{sec:sup:aux-obj}

In addition to TRM, we apply an auxiliary objective during pre-training, which aligns video features with corresponding captions by contrastive learning, widely used in image- and video-language pre-training \cite{jia2021scaling,li2021align,luo2020univl, cbt,zellersluhessel2021merlot, all_in_one_2022}. Specifically, with the concatenated video feature sequence $\mathbf{e} = \{e_1^1, ..., e_{M_1}^1,e_1^2, ..., e_{M_K}^K\}$, we add the beginning and the end token before and after the sequence, as well as the temporal position encoding to each feature. Then after contextualization, we have $\mathbf{v} = \{v_\mathtt{bos}, v_1^1, ..., v_{M_1}^1,v_1^2, ..., v_{M_K}^K, v_\mathtt{eos}\}$. To align each video to its caption, the objective learns a similarity function $\mathrm{sim}(v, c) = g_v(f_v(v))^\mathsf{T} g_c(f_c(c))$, such that parallel video-caption pairs have higher similarity scores. $f_v$ produces the representation of $\mathcal{V}_k$, which averages the features of a video, \emph{e.g.} $f_v(\mathcal{V}_k) = \sum_{m=1}^{M_k}v_m^k$, and $f_c$ delivers the representation of a caption, which is the \texttt{[CLS]} embeddings of the caption feature encoded by the question encoder. $g_v$ and $g_c$ are two linear transformations that map the two representations into a normalized lower-dimensional space.

Following \cite{li2021align}, we calculate the softmax-normalized video-to-caption and caption-to-video similarity as:
\begin{equation}
    p_k^\mathrm{v2c}(\mathcal{V}_k) = \frac{\mathrm{exp}(\mathrm{sim}(\mathcal{V}_k, \mathcal{C}_k)/\tau)}{\sum_{i=1}^K \mathrm{exp}(\mathrm{sim}(\mathcal{V}_k, \mathcal{C}_i)/\tau)},~~~~
    p_k^\mathrm{c2v}(\mathcal{C}_k) = \frac{\mathrm{exp}(\mathrm{sim}(\mathcal{C}_k, \mathcal{V}_k)/\tau)}{\sum_{i=1}^K \mathrm{exp}(\mathrm{sim}(\mathcal{C}_k, \mathcal{V}_i)/\tau)},
\end{equation}
where $\tau$ is a learnable temperature parameter. To increase the difficulty, we collect video-caption pairs from all video sequences in the same mini-batch $B$, and thus $K$ is $K$ times the size of a mini-batch in practice. Then, similar to \cite{li2021align, opai_clip}, let $\bm{y}^\mathrm{v2c}(v)$ and $\bm{y}^\mathrm{c2v}(c)$ denote the ground-truth one-hot similarity, where the probability of positive and negative pair are 1 and 0. The video-caption contrastive loss is defined as the cross-entropy CE between $\bm{p}$ and $\bm{y}$:
\begin{equation}
    \mathcal{L}_\text{align} = \frac{1}{2} \mathbb{E}_{(\mathcal{V}, \mathcal{C})\sim B} [\mathrm{CE}(\bm{y}^\mathrm{v2c}(\mathcal{V}), \bm{p}^\mathrm{v2c}(\mathcal{V})) + \mathrm{CE}(\bm{y}^\mathrm{c2v}(\mathcal{C}), \bm{p}^\mathrm{c2v}(\mathcal{C}))]
\end{equation}

\subsubsection{Pre-training Datasets}

TRM samples video-caption pairs from video captioning datasets. We want the datasets as diverse as possible, not limited to cooking \cite{youcook}, movies \cite{lsmdc}, or indoor actions \cite{charades}. To maintain the computation within an affordable size, videos cannot be too long \cite{krishna2017dense}, or a video sequence would consist of few videos, which prohibits the model from learning long-term temporal dependency.

We pre-train the video-language encoder over VATEX \cite{Wang_2019_ICCV} and TGIF \cite{tgif-cvpr2016}. VATEX contains 41K videos from Kinetics-600 \cite{kinetics} and 826K sentences, where each video is paired with multiple descriptions. The lengths of the videos are all 10 seconds, cropped for precise action recognition in Kinetics. TGIF is an open-domain dataset containing 100K animated GIFs from Tumblr and 120K sentence descriptions. The average duration is around 3.1 seconds. It is worth noting that using less pre-training data is not the main motivation of this work, but with effective objectives, our method has surpassed large-scale pre-training. If computational cost is affordable, training with more data is expected to advance the performance. We leave pre-training with longer videos and larger datasets for future work.

\begin{table}[ht]
\centering \small
    \begin{tabular}{lccc}
    \toprule
        Hyperparameter & Pre-train & ActQA & AGQA   \\
    \midrule
        Learning Rate (Base)        & 1e-5   & 2e-5  & 2e-5 \\
        Learning Rate (Video)       & 5e-5   & 2e-4  & 5e-5 \\
        Learning Rate (MLP)         & 2.5e-4 & 1e-3  & 2e-4 \\
        Learning Rate (Ans)         & 2e-5   & 2e-5  & 2e-5 \\
        Weight Decay                & 1e-2   & 1e-2  & 1e-2 \\
        AdamW $\epsilon$            & 1e-8   & 1e-8  & 1e-8 \\
        AdamW $\beta_1$             & 0.9    & 0.9   & 0.9  \\
        AdamW $\beta_2$             & 0.98   & 0.98  & 0.98 \\
        Training Steps              & 60K    & -     & -    \\
        Training Epochs             & -      & 5     & 4    \\
        Warmup                      & 0.03   & 0.1   & 0.1  \\  
        Batch Size                  & 128    & 64    & 64   \\
        Max Video Length            & 100    & 100   & 100  \\
        Max Question Length         & 50     & -     & -    \\
        Number of Videos ($K$)      & 8      & -     & -    \\
        Number of Frames ($T$)      & -      & 16    & 8    \\
    \bottomrule
    \end{tabular}
    \caption{Hyperparameters for pre-training (Pre-train), ActivityNet-QA (ActQA), and AGQA 2.0 (AGQA). Base: the question, image, and image-language encoder. Video: the video and video-language encoder. Ans: the answer encoder.}
    \label{tab:param_opt}
\end{table}
\subsection{Optimization}

The pre-training and fine-tuning are all optimized with AdamW optimizer and linear decay scheduling after warmup. All experiments are run with two NVIDIA RTX 3090s, with which the pre-training takes about 18 hours. The detailed hyperparameters are provided in Table \ref{tab:param_opt}.

\subsection{Computational Cost}

The overall computation is the sum of the IL and VL models and depends on the number of input frames $T$, video lengths, and the feature extractors. Let $R$ and $S$ denote the computation of ALBEF and Just-Ask, our method costs about $TR+S$ as we stack more Transformer layers than Just-Ask, but the two streams share the question encoder. Specifically, the frozen image encoders cost about 12 GFLOPs per frame, and the video encoder performs 40 GFLOPs per window. The other modules need 28 GFLOPs.
\section{Experimental Details}\label{sec:exps}

\subsection{Details of Temporal Modeling Analysis}

Some may question our preliminary analysis of temporal modeling, in which we first train a model with normal inputs and test it with normal and shuffled inputs. The performance drops imply the sensitivity to the order of frames, and thus little difference may indicate the incompetence of temporal modeling. Training and testing a model with shuffled input can also completely eliminate the temporal information, but this approach only reveals how well a model solves a task with spatial information (or dataset bias if the task is designed for evaluating temporal modeling), and thus it is not suitable for assessing a model's capability of temporal modeling.

We conduct the analysis on AGQA and VIOLIN as some other video QA benchmarks are less appropriate. For example, some questions in ActivietNet-QA need only spatial knowledge. In NeXT-QA \cite{xiao2021next}, while 29\% of questions are about temporal relations, others aim at spatial information or more advanced cognition, \eg causal reasoning. The split of \emph{State Transition} in TGIF-QA \cite{Jang_2017_CVPR}, though expected to suit this analysis well, could be solved by VIOLET without understanding the order of frames in our experiment (Table \ref{tab:tgif}).

\begin{table}[ht]
    \centering \small
    \begin{tabular}{ccl}
        \toprule
        Method & Benchmark & Accuracy  \\
        \midrule
        \multirow{2}{*}{VIOLET} & TGIF-QA~~ & 95.34 \\
         & TGIF-QA* & 95.36{\scriptsize$\pm$.08} \\
        \bottomrule
    \end{tabular}
    \caption{Results of VIOLET taking shuffled frames as input on the questions of \emph{State Transition} of TGIF-QA. (* signifies that input frames are shuffled. We report the average of three results for the shuffle experiment.)}
    \label{tab:tgif}
\end{table}

\subsection{Pre-training Data Used by Prior Approaches}\label{sec:dataset}

Compared with state-of-the-art approaches, DeST performs better on ActivityNet-QA with orders of magnitude less pre-training data. We include some widely-used pre-training datasets that are abbreviated in Table \ref{tab:anet} of the main paper: 100M: HowTo100M \cite{Miech_2019_ICCV}; 69M: HowToVQA69M \cite{yang2021just}; 180M:  YT-Temporal-180M \cite{zellersluhessel2021merlot}; 2.5M: WebVid \cite{Bain21}; 14M/3M: Conceptual Caption \cite{sharma2018conceptual, Changpinyo_2021_CVPR}; 5.6M: COCO \cite{chen2015coco} + VisualGenome \cite{krishna2017visual}.

\subsection{Full Results and Analysis on AGQA 2.0}\label{sec:agqa-full}

AGQA 2.0 provides extensive annotations. Each question is associated with the reasoning abilities necessary to answer the question. The annotations cover four aspects: reasoning types, semantics class, structures, and answer types. Reasoning types define the design of question templates for evaluating certain reasoning abilities. We list some examples of question templates created by \cite{GrundeMcLaughlin2021AGQA} in Table \ref{tab:agqa_type} for the following analysis of our model's behavior. The semantics class of a question describes its main subject: an object, relationship, or action. Question structures include open questions (query), comparing attributes of two options (compare), choosing between two options (choose), yes/no questions (verify), and understanding of logical operators, such as \emph{and} or \emph{or}. Questions with binary answer types have restricted answer choices, such as Yes/No, Before/After, or two specified options, while the answers to open-ended questions are much more diverse.
\begin{table}[ht]
    \centering \footnotesize
    \begin{tabular}{ll}
    \toprule
        Reasoning Type          & Example of Template                                       \\
    \midrule
        Object-Relationship     & What/Who/When/Where/How did they <rel> <object>?          \\
        Relationship-Action     & Did they <relation> something before or after <action>?   \\
        Object-Action           & Did they interact with <object> before or after <action>? \\
        Superlative             & What were they <action> first/last?                       \\
        Sequencing              & What did the person do after <action>?                    \\
        Exists                  & Did/Does/Do <concept> occur?                              \\
        Duration Comparison     & Did they <action1> or <action2> for longer?               \\
        Activity Recognition    & What does the person do before/after/while <action>?      \\
    \bottomrule
    \end{tabular}
    \caption{Reasoning types and examples of their templates of AGQA 2.0.}
    \label{tab:agqa_type}
\end{table}

\setbox0\hbox{\scriptsize \setlength{\tabcolsep}{0pt} \tabular{@{}l}Reasoning\endtabular}
\setbox1\hbox{\scriptsize \setlength{\tabcolsep}{0pt} \tabular{@{}l}Semantic\endtabular}
\setbox2\hbox{\scriptsize \setlength{\tabcolsep}{0pt} \tabular{@{}l}Structure\endtabular}
\setbox3\hbox{\scriptsize \setlength{\tabcolsep}{0pt} \tabular{@{}l}Overall\endtabular}

\begin{table}[bht]
    \centering \footnotesize
    \begin{tabularx}{0.8\linewidth}{cl *{4}{Y}}
        \toprule
        & Type                 & Just-Ask* & Just-Ask & VIOLET* & VIOLET \\
        \midrule
        \multirow{8}{*}{\rotatebox{90}{\usebox0}} 
        & Object-Relationship  & 46.30     & 47.83    & 49.01   & 48.91  \\
        & Relationship-Action  & 50.78     & 66.55    & 50.04   & 50.02  \\
        & Object-Action        & 50.77     & 68.78    & 50.13   & 50.24  \\
        & Superlative          & 37.96     & 39.83    & 39.47   & 39.49  \\
        & Sequencing           & 50.66     & 67.01    & 49.86   & 49.91  \\
        & Exists               & 57.15     & 59.35    & 54.58   & 54.70  \\
        & Duration Comparison  & 50.66     & 50.49    & 30.70   & 30.64  \\
        & Activity Recognition & 19.87     & 21.53    & ~~3.13  & ~~3.13 \\
        \midrule
        \multirow{3}{*}{\rotatebox{90}{\usebox1}} 
        & Object               & 46.34     & 49.31    & 49.18   & 49.08  \\
        & Relationship         & 54.63     & 59.60    & 52.32   & 52.41  \\
        & Action               & 49.78     & 58.03    & 41.47   & 41.45  \\
        \midrule
        \multirow{5}{*}{\rotatebox{90}{\usebox2}} 
        & Query                & 45.53     & 47.25    & 48.15   & 47.98  \\
        & Compare              & 50.84     & 65.11    & 47.65   & 47.69  \\
        & Choose               & 39.78     & 41.00    & 46.97   & 46.90  \\
        & Logic                & 54.87     & 56.20    & 50.99   & 51.24  \\
        & Verify               & 56.22     & 58.13    & 55.42   & 55.46  \\
        \midrule
        \multirow{3}{*}{\rotatebox{90}{\usebox3}} 
        & Binary               & 49.95     & 55.35    & 50.30   & 50.33  \\
        & Open                 & 45.53     & 47.25    & 48.15   & 47.98  \\
        & All                  & 47.72     & 51.27    & 49.22   & 49.15  \\
        \bottomrule
    \end{tabularx}
    \caption{Full results of the preliminary analysis of temporal modeling on AGQA 2.0. (* means shuffled input. We report the result of one experiment.)}
    \label{tab:shuffle_full}
\end{table}
\subsubsection{Full Results of Temporal Modeling Analysis}

The full results of Table \ref{tab:pre-tem} in the main paper are presented in Table \ref{tab:shuffle_full}, where we gauge the efficacy of temporal modeling of prior approaches by inputting shuffled videos and measuring performance drop. While Just-Ask \cite{yang2021just} demonstrates improvement in \emph{Relationship-Action}, \emph{Object-Action} and \emph{Sequencing}, VIOLET \cite{violet_2022} performs similar in most types. The poor performance of VIOLET may be attributed to sparsely sampling, by which they enabled end-to-end training, but it turns out that taking few frames seems not able to summarize the temporal dynamics of whole videos.

\begin{table}[h!]
    \centering \footnotesize 
    \begin{tabularx}{\linewidth}{l *{8}{Y}} 
        \toprule
        Type                 & T     & T+F   & T+\textbf{F}  & T+V    & T+\textbf{V}  & T+\textbf{F}+V & T+\textbf{F}+\textbf{V}* & T+\textbf{F}+\textbf{V} \\
        \midrule
        Object-Relationship  & 39.15 & 49.21 & 50.33 & 51.67  & 53.40 & 56.39  & 57.16   & 59.66   \\
        Relationship-Action  & 50.05 & 50.61 & 50.00 & 49.83  & 71.57 & 53.25  & 51.64   & 72.98   \\
        Object-Action        & 49.99 & 50.11 & 50.00 & 50.03  & 74.74 & 56.27  & 54.42   & 75.20   \\
        Superlative          & 34.00 & 37.96 & 38.87 & 41.82  & 43.80 & 44.54  & 45.70   & 48.94   \\
        Sequencing           & 49.89 & 50.26 & 49.86 & 49.86  & 72.60 & 54.92  & 53.14   & 73.53   \\
        Exists               & 50.09 & 57.77 & 59.06 & 50.86  & 53.68 & 59.95  & 59.04   & 63.21   \\
        Duration Comparison  & 48.71 & 51.43 & 55.04 & 44.96  & 37.34 & 62.58  & 60.26   & 60.39   \\
        Activity Recognition & 14.63 & 14.81 & 16.84 & 13.16  & 19.60 & 21.25  & 21.44   & 27.78   \\
        \midrule
        Object               & 39.25 & 49.24 & 50.16 & 51.44  & 55.28 & 56.50  & 57.31   & 61.27   \\
        Relationship         & 50.08 & 54.73 & 55.76 & 50.58  & 57.14 & 57.33  & 56.07   & 63.93   \\
        Action               & 48.49 & 49.98 & 50.86 & 47.09  & 56.52 & 56.35  & 54.39   & 65.96   \\
        \midrule
        Query                & 33.28 & 48.18 & 49.33 & 51.99  & 56.48 & 57.46  & 58.90   & 61.22   \\
        Compare              & 49.99 & 50.62 & 50.73 & 49.42  & 68.28 & 56.11  & 54.23   & 72.04   \\
        Choose               & 48.10 & 46.24 & 46.76 & 49.50  & 42.38 & 50.34  & 50.40   & 53.01   \\
        Logic                & 50.03 & 54.28 & 56.36 & 50.68  & 51.91 & 57.52  & 55.78   & 59.18   \\
        Verify               & 49.98 & 57.48 & 58.45 & 51.28  & 53.44 & 59.49  & 59.45   & 63.02   \\
        \midrule
        Binary               & 49.47 & 52.00 & 52.70 & 50.17  & 54.74 & 55.76  & 55.01   & 62.61   \\
        Open                 & 33.28 & 48.18 & 49.33 & 51.99  & 56.48 & 57.46  & 58.90   & 61.22   \\
        All                  & 41.32 & 50.07 & 51.00 & 51.08  & 55.62 & 56.61  & 56.97   & 61.91   \\
        \bottomrule
    \end{tabularx}
    \caption{Full results of our method on AGQA 2.0 with ablation of components and pre-training strategies. (T: questions; F: frames; \textbf{F}: frames with the image-language encoder pre-trained on VQA; V: videos; \textbf{V}: videos with the video-language encoder pre-trained with TRM; *: shuffled video inputs.)}
    \label{tab:agqa_full}
\end{table}
\subsubsection{Full Results and Analysis of Our Method}

We show the full results of our method on AGQA 2.0 with ablation of components and pre-training strategies in Table \ref{tab:agqa_full}.

We first examine the performance of inputting only questions (T), which reveals the bias of the datasets as these questions can be solved without grounding to videos. With a rigorous balancing procedure, this model cannot achieve more than 50\% accuracy on any question type, but some questions, for example, those belonging to \emph{Relationship-Action}, \emph{Object-Action}, and \emph{Exists} appear easier than others.

Inputting frames (T+F) improves the overall performance by about 10\% accuracy, which mostly comes from \emph{Object-Relationship} and \emph{Exists}. This is reasonable as these questions involve less temporal information according to the templates, and they are more likely to be solved with a few static frames with spatial information about humans, objects, and scenes. Pre-training the image-language encoder with VQA \cite{balanced_vqa_v2} (T+\textbf{F}) shows further improvement in \emph{Exists}, which seems more similar to the question design of image QA.

Accessing videos (T+V) is helpful for different question types such as \emph{Superlative}, in which the questions ask about something happening first or last, but some other questions that also require temporal modeling, including \emph{Relationship-Action} or \emph{Sequencing}, are not improved. Besides, video inputs do not enhance the performance of questions improved by frame inputs. This complementary advantage of frames and videos is consistent with our findings in the preliminary analysis, and inputting both frames and videos (T+\textbf{F}+V) does surpass inputting only one of them in all reasoning types.

Pre-training the video-language encoder with TRM (T+\textbf{F}+\textbf{V}) boosts the performance of most reasoning types, especially \emph{Relationship-Action}, \emph{Object-Action}, and \emph{Sequencing}. These questions all need temporal modeling of event sequences in videos and have question formats more similar to TRM. The huge performance gap (20\% accuracy) between normal (T+\textbf{F}+\textbf{V}) and shuffled video inputs (T+\textbf{F}+\textbf{V}*), as well as the little gap between no pre-training (T+\textbf{F}+V) and shuffled inputs (T+\textbf{F}+\textbf{V}*), suggests successful temporal modeling and verifies the efficacy of TRM. 

Despite the enhancement in most questions, TRM still struggles with some reasoning types, for example, \emph{Duration Comparison}, asking a machine which action lasts longer. These questions require a machine to memorize multiple events and identify their starting and ending point to obtain their duration. Such abilities are beyond the intention of developing TRM, and we leave it for future exploration.

\begin{table}[th]
\RawFloats \centering \footnotesize
\begin{minipage}{.48\textwidth} \centering
    \begin{tabular}{lrr}
        \toprule
        Type                 & VL & IL \\
        \midrule
        Object-Relationship  & 49.04          & 20.91          \\
        Relationship-Action  & 50.00          & 0.60           \\
        Object-Action        & 50.00          & 0.99           \\
        Superlative          & 36.00          & 15.86          \\
        Sequencing           & 49.85          & 0.80           \\
        Exists               & 57.09          & 0.07           \\
        Duration Comparison  & 58.99          & 0.00           \\
        Activity Recognition & 13.06          & 0.92           \\
        \midrule
        Object               & 48.92          & 20.60          \\
        Relationship         & 54.58          & 0.20           \\
        Action               & 52.19          & 0.38           \\
        \midrule
        Query                & 47.35          & 26.68          \\
        Compare              & 51.24          & 0.69           \\
        Choose               & 48.29          & 22.11          \\
        Logic                & 55.09          & 0.04           \\
        Verify               & 56.51          & 0.08           \\
        \midrule
        Binary               & 52.52          & 6.30           \\
        Open                 & 47.35          & 26.68          \\
        All                  & 49.91          & 16.56          \\  
        \bottomrule
    \end{tabular}
    \vspace{16pt}
    \captionof{table}{Full results of the ablation study on two encoding streams. (VL: ablating the video-language encoder; IL: ablating the image-language encoder.)}
    \label{tab:ensem_full}
\end{minipage} \hspace{0.02\textwidth}
\begin{minipage}{.47\textwidth}\centering
    \begin{tabular}{cccc}
        \toprule
        Question & Frames & Video & Acc \\
        \midrule
        $\checkmark$ &         &         & 31.01 \\
        $\checkmark$ & $\checkmark$ &    & 43.15 \\
        $\checkmark$ & VQA     &         & 46.66 \\
        \midrule
        $\checkmark$ & VQA     & TRM     & 46.79 \\
        \bottomrule
    \end{tabular}
    \vspace{16pt}
    \captionof{table}{Ablation study on ActivityNet-QA. ($\checkmark$ means the modality is presented. VQA: pre-trained on VQA. TRM: pre-trained with TRM.)}
    \label{tab:ablate_anet}
\end{minipage}
\end{table}
\subsubsection{Full Results of Ablation Study of Encoding Streams}

Table \ref{tab:not-ensem} in the main paper is expanded as Table \ref{tab:ensem_full}, where we first train a model with both image- and video-language encoders, and evaluate each stream with the test set. 

\subsection{Ablation Study on ActivityNet-QA}

The ablation study is also conducted on ActivityNet-QA, as reported in Table \ref{tab:ablate_anet}. The result proves that the image-language model is capable of answering some video QA problems, and the strategy of bridging image QA and video QA further increases the performance.

\subsection{Temporal Resolutions of the Image-Language Encoder}

\begin{figure}[h]
\centering
\begin{minipage}{0.45\textwidth}
    \centering
    \includegraphics[width=\textwidth]{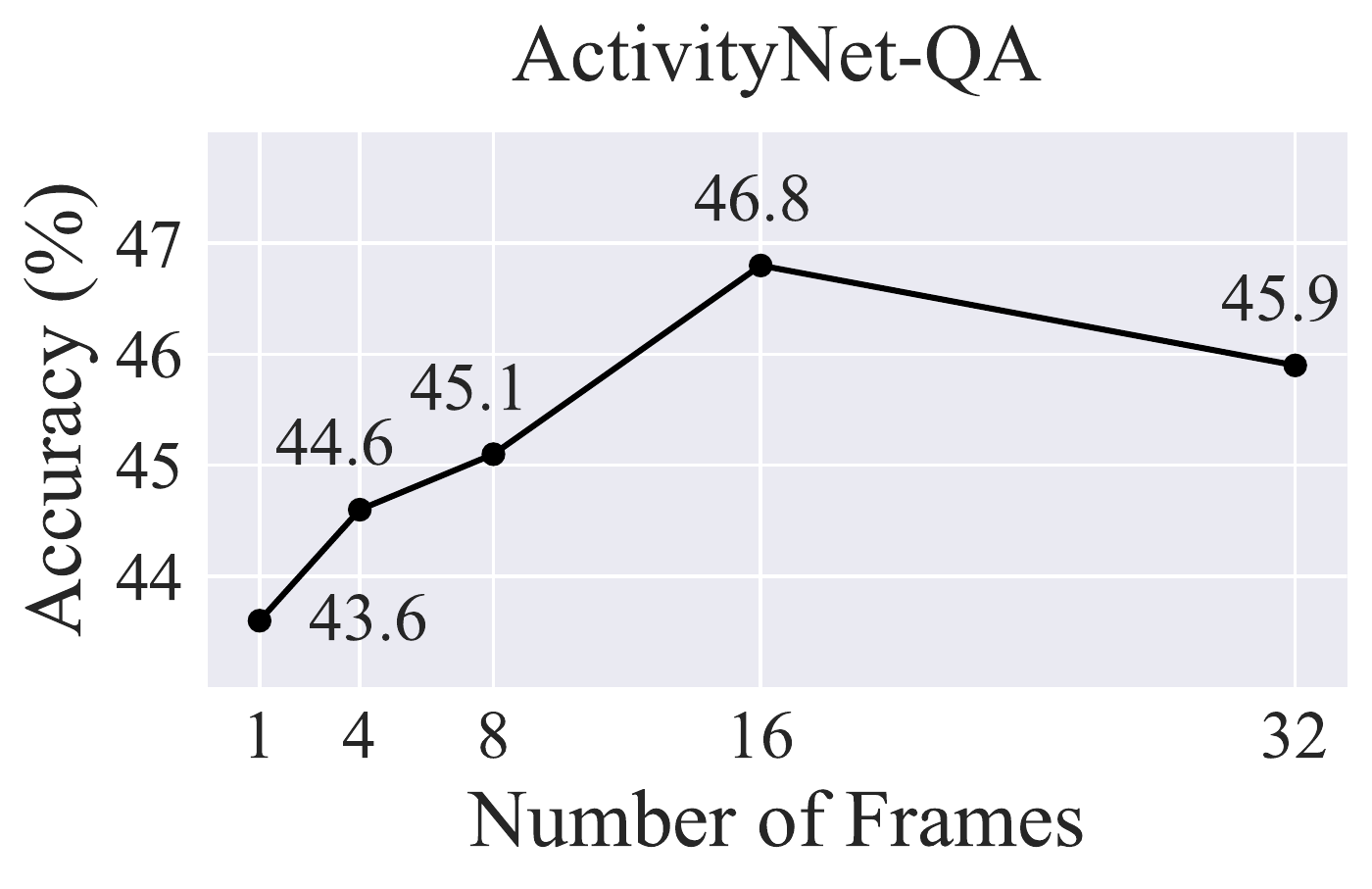}
\end{minipage} \hspace{0.05\textwidth}
\begin{minipage}{0.45\textwidth}
    \centering
    \includegraphics[width=\textwidth]{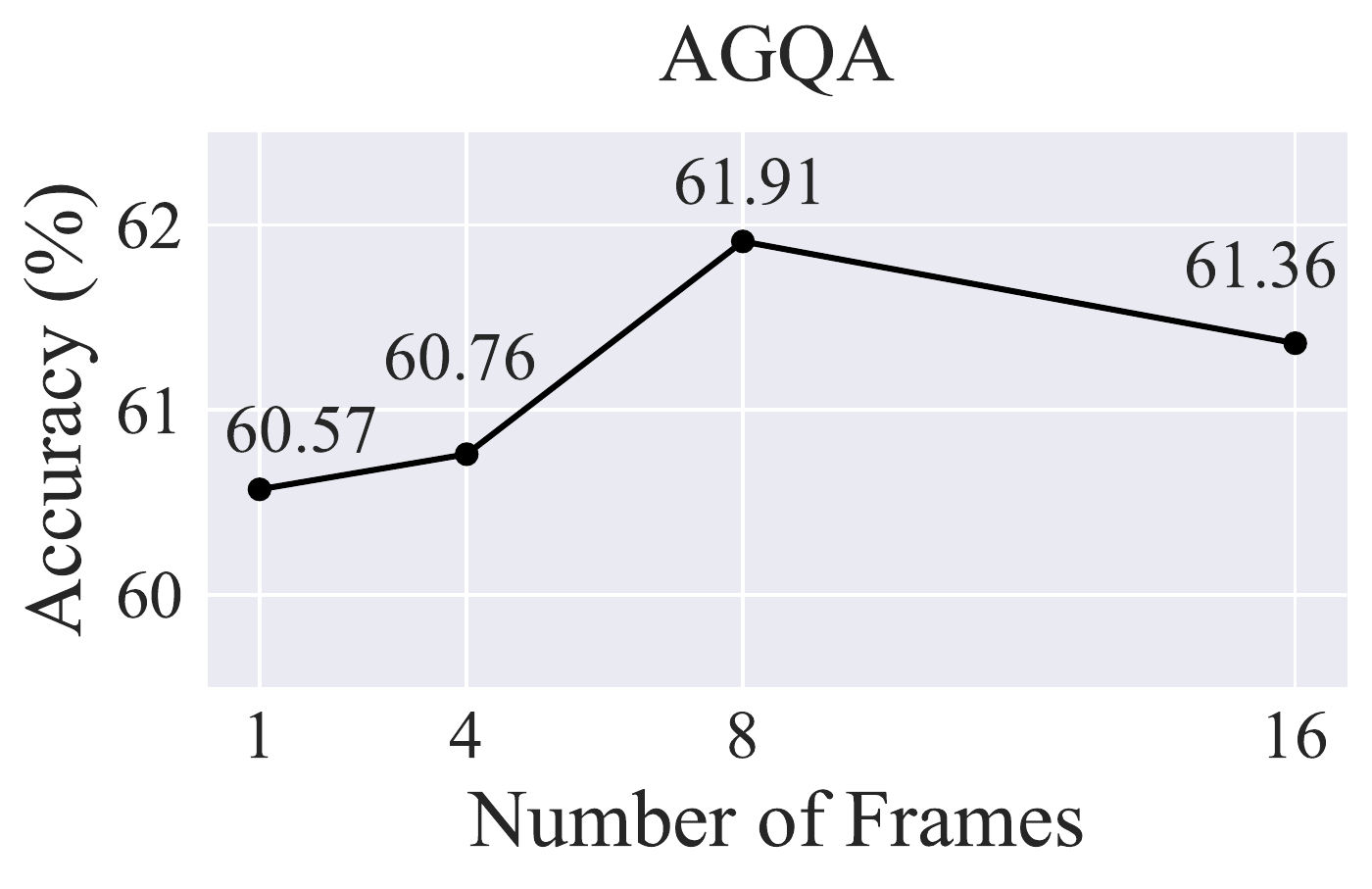}
\end{minipage}
\caption{The results on two benchmarks of inputting different numbers of frames to the image-language encoder.}
\label{fig:nof}
\end{figure}

We estimate the influence of varying temporal resolutions (the number of frames $T$) of the image-language encoder. As displayed in Figure \ref{fig:nof}, taking more frames substantially increases the performance on ActivityNet-QA, while the improvement on AGQA is insignificant. This discrepancy could be explained by the distribution of question types in the two benchmarks, where ActivityNet-QA contains more questions of static information and the prediction is likely stabler and more robust when more frames are provided. More questions in AGQA are related to temporal dynamics and thus less affected by the number of frames.

\subsection{The Number of Videos for Temporal Referring Modeling}

\begin{figure}[ht]
    \centering
    \includegraphics[width=.45\textwidth]{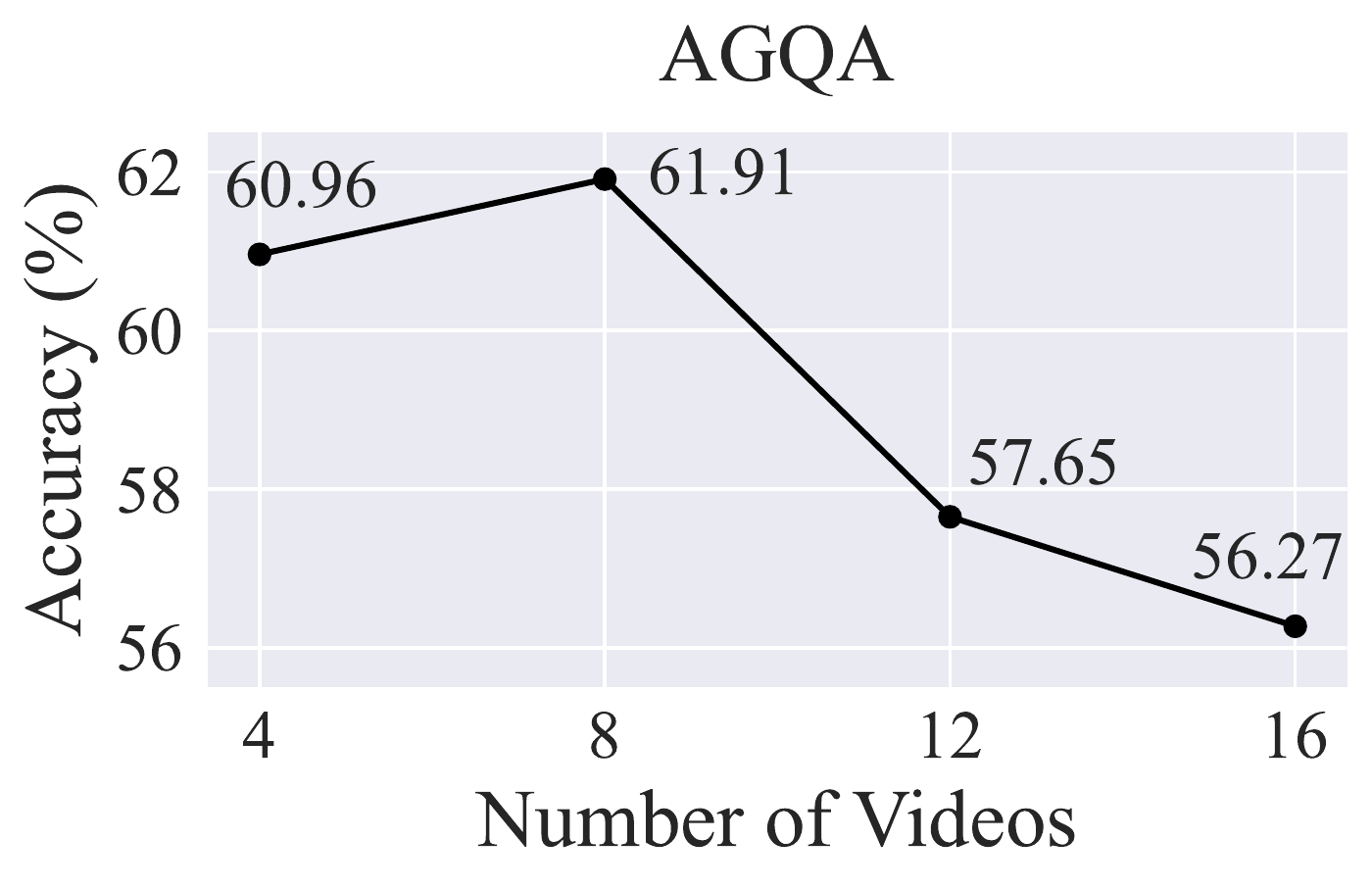}
    \caption{The performance on AGQA 2.0 of varying the numbers of concatenated videos for Temporal Referring Modeling.}
    \label{fig:nov}
\end{figure}

We alter the number of videos concatenated for TRM (the variable $K$) and study its influence. The accuracy on AGQA 2.0 with respect to the number of videos is presented in Figure \ref{fig:nov}. We can observe that increasing the number of videos is not always beneficial to the downstream task. Concatenating too many videos may result in extremely long temporal dependency, which is hard for a model to encode.

\subsection{Multi-Task Loss Weighing}

As described in \ref{sec:sup:aux-obj}, we align visual and linguistic content by contrastive learning. An experiment is conducted to evaluate different approaches to loss combinations. Let $\mathcal{L}_\text{TRM}$ and $\mathcal{L}_\text{align}$ denote the loss of TRM and video-language contrastive loss. We compare adding losses directly $\mathcal{L}_1$ and weighing losses by uncertainty \cite{Kendall_2018_CVPR} $\mathcal{L}_2$:
\begin{equation}
\begin{split}
    &    \mathcal{L}_1 = \mathcal{L}_\text{TRM} + \mathcal{L}_\text{align}, \\
    &    \mathcal{L}_2 = \frac{1}{2\sigma_1^2}\mathcal{L}_\text{TRM} + \frac{1}{2\sigma_2^2}\mathcal{L}_\text{align} + \log \sigma_1^2 + \log \sigma_2^2,
\end{split}
\end{equation}
where $\sigma_1$ and $\sigma_2$ are two learnable parameters. Following \cite{Kendall_2018_CVPR}, in practice the model learns to predict $s := \log \sigma^2$ for numerical stability.

\begin{table}[th]
    \centering \footnotesize
    \begin{tabular}{lc}
        \toprule
        Loss combination         & Acc   \\
        \midrule
        Unweighted               & 61.91 \\
        Weighing by uncertainty  & 60.15 \\
        \bottomrule
    \end{tabular}
    \caption{Comparison between different approaches to loss combination.}
    \label{tab:loss-combine}
\end{table}

Table \ref{tab:loss-combine} compares the accuracy on AGQA 2.0 of pre-training with the sum of losses unweighted and weighted by uncertainty. The result shows that the difference between the two approaches to combining losses is insignificant.

\end{document}